\pdfoutput=1

\documentclass[11pt]{article}

\usepackage{acl}

\usepackage{times}
\usepackage{latexsym}
\usepackage{inconsolata}
\usepackage{amsfonts}
\usepackage{soul}
\usepackage{colortbl}
\usepackage{multirow}
\usepackage{xspace}
\usepackage{amsmath}
\usepackage{graphicx}
\usepackage{booktabs}
\usepackage{tcolorbox}
\usepackage{caption}
\usepackage{subcaption}
\usepackage{wrapfig}
\usepackage{enumitem}

\newcommand{\FT}{ERM\xspace}
\newcommand{\KW}{KernelWhitening\xspace}
\newcommand{\ETE}{HypothesisOnly-PoE\xspace}
\newcommand{\WL}{WeakLearner-PoE\xspace}

\newcommand{\READ}{AttentionPoE\xspace}
\newcommand{\DAMP}{$\sigma$-Damp\xspace}
\newcommand{\DFR}{DeepFeatReweight\xspace}
\newcommand{\PGD}{PerSampleGrad\xspace}

\newcommand{\teascale}{$\mathcal{T}$\xspace}
\newcommand{\stuscale}{$\mathcal{S}$\xspace}
\newcommand{\teacher}{$f_{\mathcal{T}}$\xspace}
\newcommand{\studentnonkd}{$f_{\mathcal{S}}$\xspace}
\newcommand{\studentkd}{$g_{{\mathcal{T}}->\mathcal{S}}$\xspace}

\usepackage[T1]{fontenc}

\usepackage[utf8]{inputenc}

\usepackage{microtype}

\usepackage{inconsolata}

\title{Do Students Debias Like Teachers?\\On the Distillability of Bias Mitigation Methods}

\author{Jiali Cheng \\
  University of Massachussetts Lowell \\
  \texttt{jiali\_cheng@uml.edu} \\\And
  Chirag Agarwal \\
  University of Virginia \\
  \texttt{chiragagarwal@virginia.edu} \\\AND
  Hadi Amiri \\
  University of Massachussetts Lowell \\
  \texttt{hadi\_amiri@uml.edu} \\
  }

\begin{document}
\maketitle

\begin{abstract}

Knowledge distillation (KD) is an effective method for model compression and transferring knowledge between models. However, its effect on model's robustness against spurious correlations that degrade performance on out-of-distribution data remains underexplored. This study investigates the effect of knowledge distillation on the transferability of ``debiasing'' capabilities from teacher models to student models on natural language inference (NLI) and image classification tasks. 
Through extensive experiments, we illustrate several key findings: 
(i) overall the debiasing capability of a model is undermined post-KD; 
%
(ii) training a debiased model does not benefit from injecting teacher knowledge;
(iii) although the overall robustness of a model may remain stable post-distillation, significant variations can occur across different types of biases; and 
(iv) we pin-point the internal attention pattern and circuit that causes the distinct behavior post-KD.
Given the above findings, we propose three effective solutions to improve the distillability of debiasing methods: developing high quality data for augmentation, implementing iterative knowledge distillation, and initializing student models with weights obtained from teacher models.
To the best of our knowledge, this is the first study on the effect of KD on debiasing and its interenal mechanism at scale. Our findings provide understandings on how KD works and how to design better debiasing methods.
\end{abstract}
\section{Introduction}\label{sec:intro}
Machine learning models are susceptible to biases or spurious correlations in datasets, commonly known to as ``shortcuts'' or ``dataset biases''~\citep{celeba,mccoy-etal-2019-right}.
Models that rely on shortcuts can achieve high performance on in-domain or over-represented data, but degrade significantly on out-of-distribution or under-represented data~\citep{Li_2023_CVPR,chew-etal-2024-understanding,Li_2025_CVPR}.

Despite recent advancements in bias mitigation~\citep{guo-etal-2023-debias,Noohdani_2024_CVPR,cheng-amiri-2024-fairflow,bombari2025spurious} and knowledge distillation~\citep{stanton2021does,sultan-2023-knowledge,Sun_2024_CVPR,he2025dakd}, the effect of knowledge distillation on debiasing at scale is largely unexplored. The internal mechanisms causing that effect remains unclear. This work studies the following research questions (RQs):
\begin{itemize}[leftmargin=*,labelsep=0.75em]
    \item \textbf{RQ1}: \textit{To what extent can knowledge distillation transfer debiasing capabilities between models?}
    
    \item \textbf{RQ2}: \textit{Can knowledge distillation train less biased models compared to standard training?} 
    
    \item \textbf{RQ3}: \textit{What internal mechanisms cause the debiasing behavior change after distillation?}
\end{itemize}

Answering these questions will help us understand the efficacy of knowledge distillation in handling dataset biases, its underlying mechanisms, and its role in developing new training methods for bias mitigation at scale.

We answer these questions by designing and conducting an empirical analysis on natural language understanding and image classification tasks. Our analyses show that:
(i) the effect of knowledge distillation on debiasing performance depends on the underlying debiasing method, the relative scale of the models involved, and the size of the training set; 
(ii) knowledge distillation effectively transfers debiasing capabilities when teacher and student are similar in scale (number of parameters); 
(iii) knowledge distillation may amplify the student model's reliance on spurious features, and this effect does not diminish as the teacher model scales up; and 
(iv) although the overall robustness of a model may remain stable post-distillation, significant variations can occur across different types of biases; and 
%
%
(v) consistent transfer patterns sometimes emerge, such as performance gap between teacher and student on out-of-distribution (OOD) data, 
suggesting the possibility of predictable changes in robustness after distillation. 
Given the above findings, we propose three effective solutions to improve the distillability of debiasing methods: developing high quality data for augmentation, implementing iterative knowledge distillation, and initializing student models with weights obtained from teacher models.\looseness-1

We summarize our contributions as follows: 
\vspace{-1mm}
\begin{itemize}
\itemsep-1pt
    \item we present the first study (to the best of our knowledge) of the effect of knowledge distillation on dataset bias at scale across both language and vision tasks;
    \item we investigate the internal mechanisms that causes debiasing ability change before and after distillation, namely the divergence of attention and change of circuit;
    \item we propose three strategies to improve the distillability of debiasing methods and provide insights for future development of bias mitigation techniques.
\end{itemize}




\section{Knowledge Distillation and Debiasing}
\paragraph{Problem Formulation}
We investigate the effect of knowledge distillation (KD) on debiasing methods. We define \emph{distillability of debiasing methods} as the amount of performance maintained before and after distilling a debiased model. We define \emph{contribution of KD} as the performance improvement gained by training a debiasing method with KD over training without KD.

\begin{figure*}[t]
    \centering
    \includegraphics[width=0.9\textwidth]{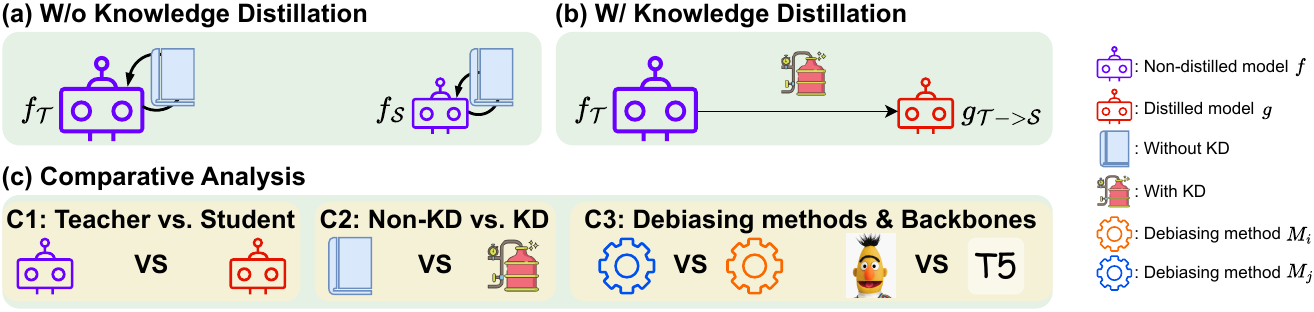}
    \caption{Framework for the analysis of distillability of debiasing methods. \textbf{(a) training from scratch}: we train a debiasing method $M_i$ from scratch without knowledge distillation on different scales (teacher \teascale and student \stuscale such that \teascale $>$ \stuscale) to obtain models \teacher, \studentnonkd respectively. 
    \textbf{(b) Training with knowledge distillation}: we apply knowledge distillation to transfer knowledge from teacher (\teacher) to student (\studentkd). 
    \textbf{(c) Assessment}: \emph{C1} determines if knowledge distillation can transfer the debiasing capability from teacher (\teacher) to student (\studentkd), 
    \emph{C2} determines the contribution of knowledge distillation in training a debiased model, and 
    \emph{C3} compares different debiasing methods and backbones under knowledge distillation.}
    \label{fig:model}
\end{figure*}

\paragraph{Notation and Training Setup}
Let $f$ and $g$ denote models trained without knowledge distillation and with knowledge distillation respectively. In this paper, we use subscript \teascale and \stuscale to denote teacher and student scales respectively. As illustrated in Figure~\ref{fig:model}, we train the following models for each debiasing method:
\textbf{(i)} we train from scratch 
for both teacher and student scales to obtain \teacher and \studentnonkd, see Figure~\ref{fig:model}(a). 
\textbf{(ii)} Then for every scale \teascale $>$ \stuscale, we distill the knowledge from \teacher to \studentkd, see Figure~\ref{fig:model}(b). 
%
Given a debiasing method $M$ and the three models obtained above (\teacher, \studentnonkd, and \studentkd), we conduct the following comparisons: 
\vspace{-1mm}
\begin{itemize}
\itemsep-1pt
    \item {\bf C1: Teacher (\teacher) vs. Student (\studentkd)}. This comparison reveals if knowledge distillation can distill debiasing capability between models and if it affects model's robustness to spurious correlations, which answers \emph{RQ1} (\S\ref{sec:c1}).

    \item {\bf C2: Non-KD vs. KD}, realized by comparing \studentnonkd vs. \studentkd. This comparison demonstrates if training bias mitigation networks can benefit from external knowledge from teacher models, which answers \emph{RQ2} (\S\ref{sec:c2}).


\end{itemize}

We note that when \teascale = \stuscale, C1 and C2 are essentially the same comparison. To avoid duplicate discussion, we will present results when \teascale = \stuscale in C2.

\section{Experimental Setup}
For consistency and fair comparison with previous debiasing works in NLU~\citep{jeon-etal-2023-improving,reif-schwartz-2023-fighting} and image classification~\citep{kirichenko2023last,last_layer_fewer,Li_2023_CVPR}, we adopt commonly used experimental setups, including choice of backbone models, datasets, evaluation protocols, and debiasing methods. In addition, all experiments are repeated three times with different random seeds to account for any stochastic effect.\looseness-1

\paragraph{Backbones}
We conduct experiments on a series of BERT~\citep{devlin-etal-2019-bert,turc2019wellread}, T5~\citep{tay2022scale}, ResNet~\citep{He_2016_CVPR}, and ViT~\citep{dosovitskiy2021an} backbones of different scales, shown in Appendix~\ref{sec:implementation} Table~\ref{tab:backbone}. These backbones are chosen for several reasons: 
BERT and ResNet are commonly employed in prior works, which enables consistent comparisons. 
In addition, ViT and T5 are commonly used backbones for vision and language tasks, but relatively less experimented in prior debiasing works, which allows investigating the generalizability of our findings beyond existing research. 
Finally, each backbone is associated with a series of publicly available pre-trained checkpoints of different scales, with consistent network architecture and pre-training data, which enables cross-scale distillation and comparisons.


\paragraph{Evaluation}
To provide a comprehensive evaluation of robustness against spurious correlations, we compare the teacher \teacher and the student \studentkd from the following perspectives:

\begin{itemize}
\itemsep0pt
    \item {\bf In-domain performance (ID, $\uparrow$)}: the performance on in-domain test set. 
    A robust model should achieve high performance on this set to demonstrate general capability.
    
    \item {\bf Out-of-domain performance (OOD, $\uparrow$)}: the performance on in-domain test set. Such samples require real task-related signals to predict, where biased models fall short. For text datasets, we evaluate models on separate OOD test sets, comprised with specially crafted hard samples~\citep{mccoy-etal-2019-right}. For image datasets, samples are divided into groups based on their labels and spurious attributes, where OOD refers to the worst performing sub-group~\citep{subpopulation}.
    
    \item {\bf Spurious gap (Spu. Gap, $\downarrow$)}: the performance gap between ID and OOD, which quantifies a model's vulnerability to spurious correlations. Ideally, a robust model should have high performances on both ID and OOD with a small spurious gap.\looseness-1 

\end{itemize}
Similarly, we compare KD and Non-KD as above. We compute F1 score on QQP and accuracy on other datasets.\looseness-1

\paragraph{Investigating Internal Mechanism}
Besides the above superficial performance metrics, we aim to uncover the internal mechanisms that causes the change of debiasing ability post-KD.
\begin{itemize}
    \item \textbf{Activation Pattern}. We conduct activation-level analysis and comparisons across layers. We use Centralized Kernel Alignment (CKA), a commonly adopted technique to measure the similarity between activation matrices or hidden representations of neural networks~\citep{cka,cka2}. Following previous work~\citep{do-vit-see-like-cnn,nguyen2021do}, we use CKA by first probing the intermediate representations from each layer and then comparing all pairwise similarities between representations of the teacher and student models, under linear kernel CKA. 

    \item \textbf{Circuit Discovery}. We also analyze and compare bias-specific sub-networks, or ``circuits''. This approach moves beyond simply observing a model's outputs to causally trace how information flows through and is processed by a coordinated set of components. Specifically, we use EAP~\citep{hanna2024have}, a widely adopted method for circuit discovery.

\end{itemize}




\paragraph{Datasets}
We use the following datasets: 1) CelebA~\citep{celeba}, 2) Waterbird~\citep{SagawaDistributionally}, 3) MNLI~\citep{williams-etal-2018-broad}, and 4) QQP~\citep{sharma2019natural}. More details of dataset statistics, causal and spurious features, and the OOD test sets are discussed in Appendix~\ref{sec:dataset}.

\paragraph{Debiasing Methods}
Experiments are conducted on a comprehensive list of commonly used debiasing methods, each of which is designed with special formulation and assumptions. We use 
(a) Empirical Risk Minimization (ERM) (standard training without debiasing techniques, 
(b) \ETE~\citep{karimi-mahabadi-etal-2020-end}, 
(c) \WL~\citep{sanh2020learning}, 
(d) \KW~\citep{gao-etal-2022-kernel}, 
(e) \READ~\citep{wang-etal-2023-robust}, 
(f) CurriculumDebiasing~\citep{lee-etal-2025-curriculum},
(g) \DAMP~\citep{NEURIPS2023_e3546030}, 
(h) \DFR~\citep{kirichenko2023last}, and 
(i) \PGD~\citep{ahn2023mitigating}. The above debiasing methods have a wide coverage of existing algorithms, ranging from auxiliary biased model-based debiasing, to disentanglement of representations. Meanwhile, they can handle multiple types of shortcuts at the same time, without overfitting to a specific bias. Details of these methods are provided in Appendix~\ref{sec:methods}. 

\begin{figure*}[t]
    \centering
    \begin{subfigure}[t]{0.49\textwidth}
    \centering
    \includegraphics[width=\textwidth]{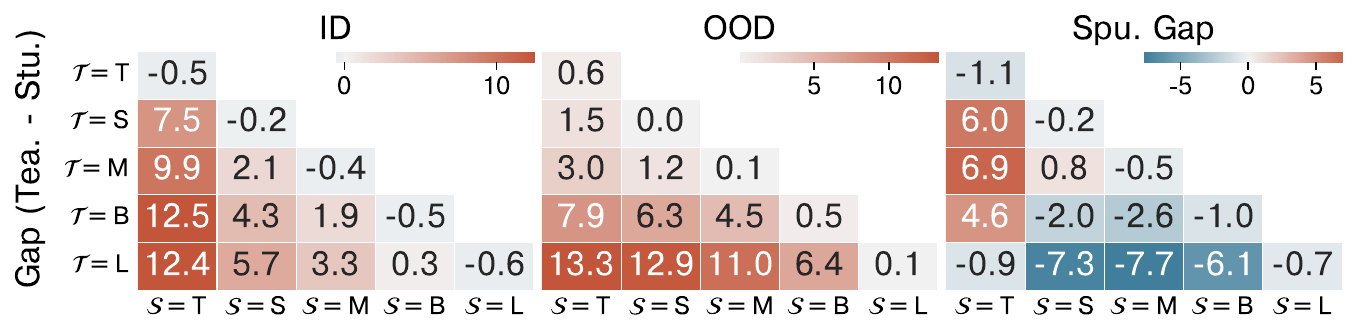}
    \caption{C1: Teacher vs. Student}
    \label{fig:heatmap_tvs}
    \end{subfigure}
    \hfill
    \begin{subfigure}[t]{0.49\textwidth}
    \centering
        \includegraphics[width=\textwidth]{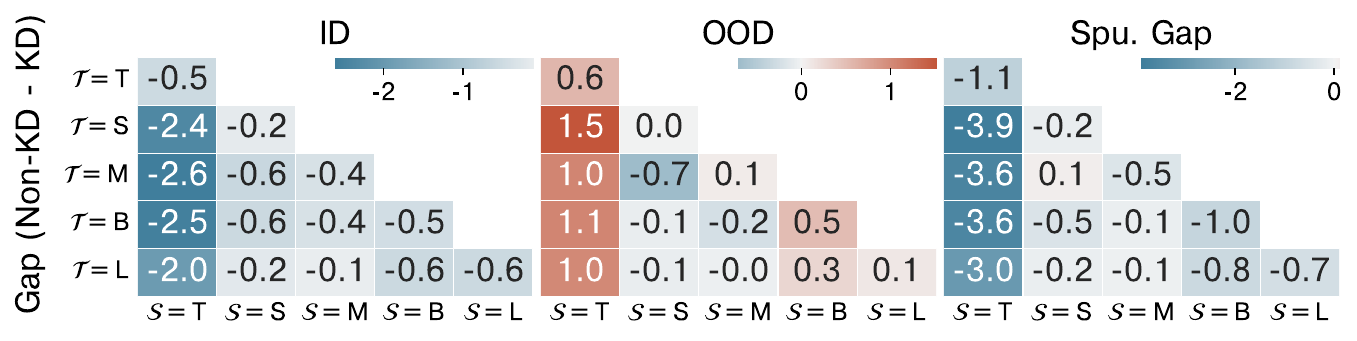}
    \caption{C2: Non-KD vs. KD}
    \label{fig:heatmap_kvn}
    \end{subfigure}
    \caption{\textbf{Average performance gaps} on ID, OOD, and Spurious Gap between (a) Teacher and Student and (b) Non-KD and KD. X-axis and Y-axis show the scale of student (\stuscale) and teacher (\teascale) respectively. Each cell shows the performance gap. See Appendix~\ref{sec:additional_results} for detailed results.}
    \label{fig:heatmap}
\end{figure*}

\section{Effect of Knowledge Distillation on Debiasing}

We first examine if KD can effectively distill the debiasing capability from teachers to students of different scales in Section~\ref{sec:c1}.
We then asses if training with knowledge distillation (KD) can improve a model's debiasing performance compared to standard training (Non-KD) in Section~\ref{sec:c2}. 
Finally, we assess the effect of different debiasing methods and backbones on our earlier findings in Section~\ref{sec:c3}.

\subsection{RQ1: Distillability of Debiasing Methods} \label{sec:c1}

\paragraph{Students become more biased than teachers}
We observe that teachers consistently achieve better performance than their smaller scale students on ID and OOD test sets after knowledge distillation. The positive values 
in Figure~\ref{fig:heatmap_tvs} show that 
although KD encourages students to mimic their teachers in the logit space, it may undesirably increase student's susceptibility to spurious correlations in datasets as the teacher's in-domain and debiasing capabilities are not effectively transferred to the student.
The prediction agreement between teacher and student models show similar trend, where the student generally aligns with the teacher on ID but often largely diverges from the teacher on OOD.
Furthermore, the extent of knowledge loss after distillation varies depending on the relative scales of the teacher and student models. 
For example, as depicted in Figure~\ref{fig:heatmap_tvs}, when \stuscale is tiny (\stuscale = T), more debiasing power is lost, shown by mostly positive values in spurious gap. When \teascale is large (\teascale = L), more ID knowledge is lost, shown by mostly negative values in spurious gap.
The results show that if a teacher model learns a partially debiased representation but still retains residual biases, the student might amplify this bias rather than mitigate it.

\paragraph{Students show diverse distribution shifts in predicted probabilities}
To understand the influence of KD on the debiasing capabilities of students, we investigate the output probability distribution $P_{\mathcal{C}}(y=1)$. Our findings show that KD significantly alter the predicted probability distribution, despite its training objective of matching output logits. This perturbation  is often larger on OOD than ID test sets, which explains the larger performance drop observed in students compared to their teachers on OOD, as illustrated in Figure~\ref{fig:density_tvs}. 
We also observed that teachers tend to provide slightly more confident predictions on ID while more moderate predictions on OOD. Such behavior is not successfully transferred to students through KD.
Such distinct behaviors on different samples may encourage models to overfit to data distributions of the training sets or to over-represented groups, which can effectively amplify reliance on shortcuts over robust features. In addition, the training sets of teachers often contain biased examples or do not equally represent all sub-groups, which leads students to inherit and potentially amplify these biases. 
Consequently, students often perform worse than their teachers on OOD.

\begin{figure*}[t]
    \centering
    \begin{subfigure}[t]{0.49\textwidth}
    \centering
    \includegraphics[width=\textwidth]{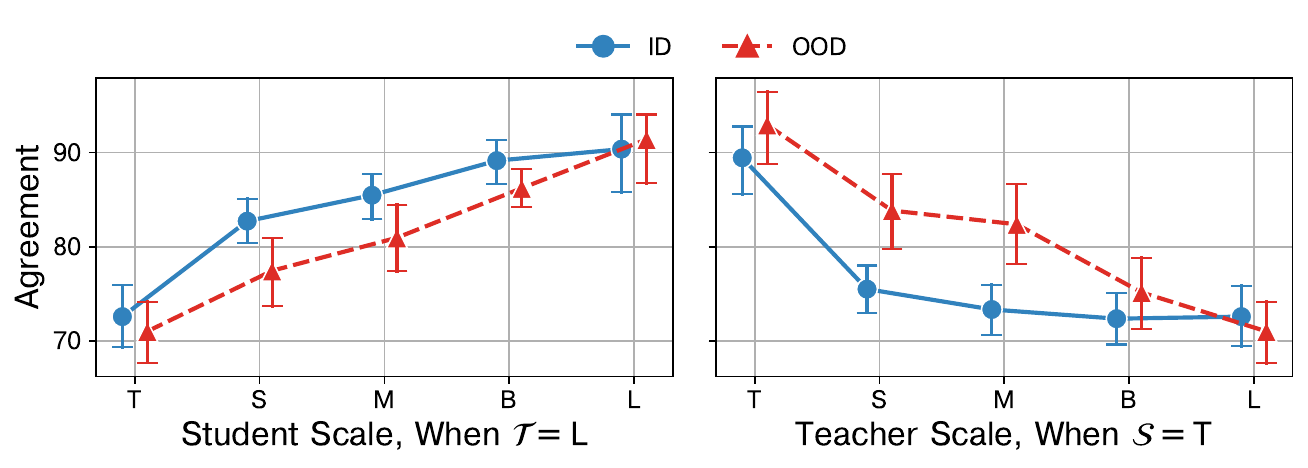}
    \caption{C1: Teacher vs. Student}
    \end{subfigure}
    \hfill
    \begin{subfigure}[t]{0.49\textwidth}
    \centering
        \includegraphics[width=\textwidth]{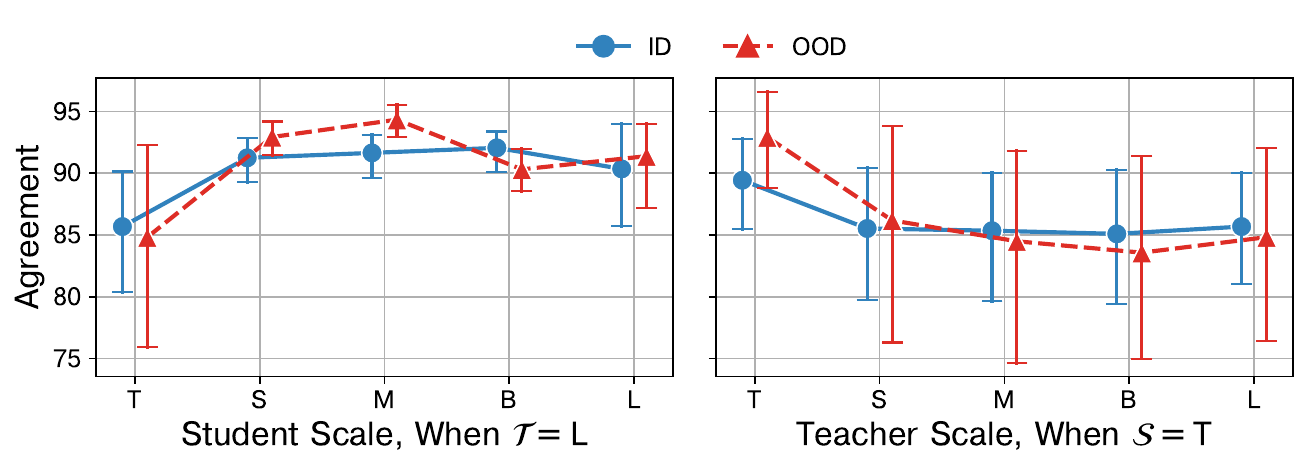}
    \caption{C2: Non-KD vs. KD}
    \end{subfigure}
    \caption{\textbf{Prediction agreement on text datasets}. 
    Agreement increases as the scale of teacher and student get closer. See Appendix~\ref{sec:additional_results} for detailed results.}
    \label{fig:agreement}
\end{figure*}

\begin{figure*}[t]
    \centering
    \begin{subfigure}[t]{0.49\textwidth}
    \centering
    \includegraphics[width=\textwidth]{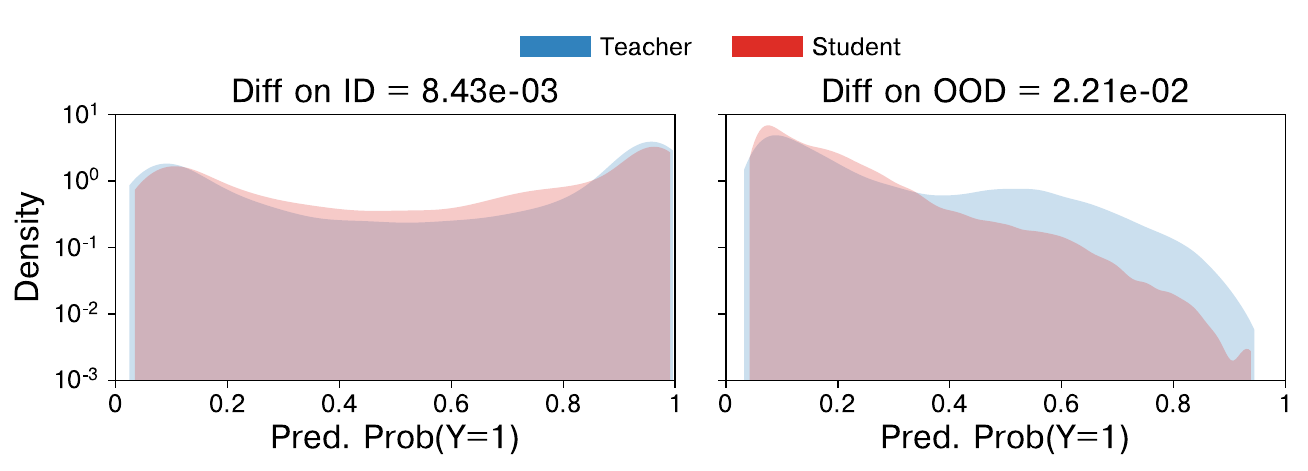}
    \caption{C1: Teacher vs. Student}
    \end{subfigure}
    \hfill
    \begin{subfigure}[t]{0.49\textwidth}
    \centering
        \includegraphics[width=\textwidth]{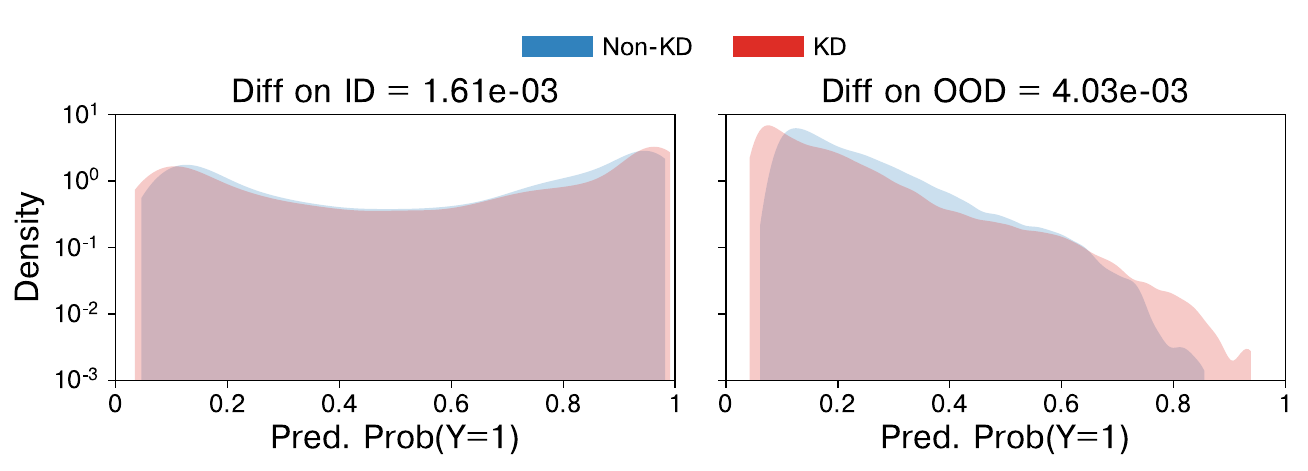}
    \caption{C2: Non-KD vs. KD}
    \end{subfigure}
    \caption{\textbf{Density of predicted probability}. On OOD, students has larger deviation in prediction confidence than teachers. See Appendix~\ref{sec:additional_results} for detailed results.}
    \label{fig:density}
\end{figure*}



\begin{figure}[t]
    \centering
    \includegraphics[width=0.49\textwidth]{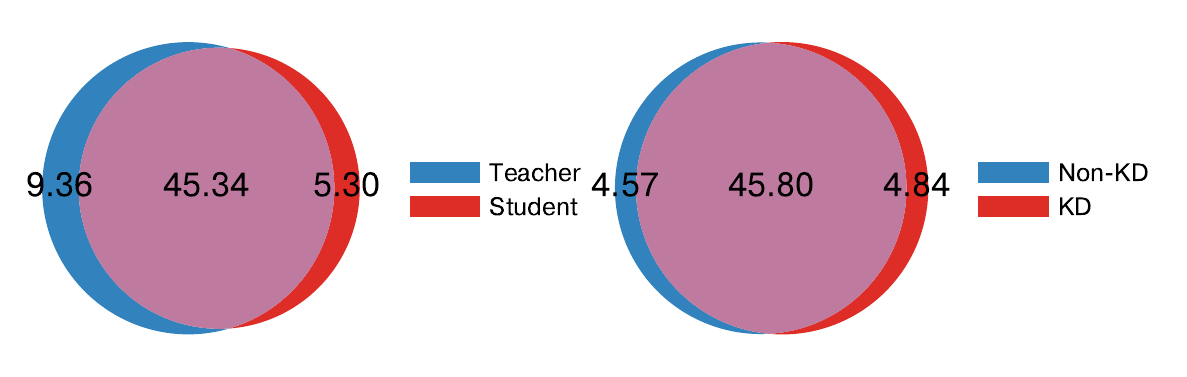}
    \caption{\textbf{C1: Teacher vs. Student} (Left) and \textbf{C2: Non-KD vs. KD} (Right): correctly predicted examples on OOD on text datasets.}
    \label{fig:venn}
\end{figure}

\paragraph{Potential for new debiasing capabilities for students beyond teacher abilities}
We compare prediction agreement 
between teacher and student models. When \teascale is large (\teascale = L), we observe an increase in prediction agreement as \stuscale scales up, with consistently higher agreement on ID than OOD, as shown in the left plot in Figure~\ref{fig:agreement}. Conversely, when \stuscale is tiny (\stuscale = T), the prediction agreement diminishes as \teascale scales up, with higher agreement on OOD than ID, the right plot in Figure~\ref{fig:agreement}. The imperfect agreement between teacher and student contradicts with the foundational assumptions of knowledge distillation, which assumes that students should closely mimic their teachers. 
However, interestingly, this unexpected behavior may not always lead to performance degradation. Sometimes it enables students to generalize to out of domain data. In particular, there are instances where students make correct predictions where their teachers do not, see the left plot in Figure~\ref{fig:venn}.
Students can sometimes outperform their teachers perhaps because they may learn additional patterns during the knowledge distillation process, which allows them to generalize better than their teachers. The above result suggests that students may sometimes acquire debiasing capabilities that surpass those of their teachers, which we believe is a novel avenue for robust model training.


\paragraph{Larger teachers do not guarantee more robust students}
Our findings show that a more capable teacher does not guarantee a less biased student in debiasing tasks. With a fixed student scale (as seen in the columns of Figure~\ref{fig:heatmap}), increasing the teacher's scale does not consistently reduce performance gap or spurious gap. Sometimes, a larger teacher may degrade the debiasing capability of the student. For example, when \stuscale = T, increasing the teacher scale from M to B increases the spurious gap from 6.5 to 8.1 on \FT, i.e. a more biased model. Moreover, when when \stuscale = T, increasing \teascale result in a drop of teacher-student agreement, indicating that the student fails to follow the teacher, see right plot in Figure~\ref{fig:agreement}. 
We attribute this finding to two reasons. Firstly, the capabilities of students are substantially bounded by their scale, and using a much larger and capable teacher may exceed the student's capacity for effective learning
~\citep{Cho_2019_ICCV}. Secondly, training students with debiasing objectives and knowledge distillation at the same time results in optimization problem, which may trap students' parameters in local optima and affect their robustness to spurious correlations.

\paragraph{Students with similar scales to their teachers learn better}
The effectiveness of debiasing ability transfer through distillation is greatly affected by the scale similarity between teacher and student. As the teacher and the student become similar in scale (near the diagonal cells in Figure~\ref{fig:heatmap}), the differences on test set performance and spurious gaps decrease. However, a larger mismatch in scale (far from diagonal) results in more pronounced differences, see Figure~\ref{fig:heatmap}. Similarly, the teacher-student agreement increases as \teascale and \stuscale align more closely, see Figure~\ref{fig:agreement}. 
This is likely because models of similar scales have comparable expressive power and extracts similar features, which can lead to more effective knowledge transfer, better bias mitigation, and higher prediction agreement.

\subsection{RQ2: Distillation vs. Standard Training} \label{sec:c2}

\paragraph{Non-KD is less biased than KD}
Our results show that debiasing models trained from scratch (Non-KD) have lower ID performance than those trained with KD. However, the Non-KD models achieve almost no changes on OOD, leading to smaller spurious gaps, see Figure~\ref{fig:heatmap}. 
We hypothesize that the distillation objective of matching logits, despite effective on ID, may potentially inject additional spurious correlations and distract the model from prioritizing robust features, as the teacher is not fully unbiased.

\paragraph{KD does not improve generalization}
An interesting finding is that both Non-KD and KD have similar average prediction agreements on both ID and OOD. However, the agreement on OOD varies significantly depending on dataset, debiasing method, and backbone model. This suggests that training solely with the original data (Non-KD) is sufficiently effective for debiasing, and introducing external knowledge via KD does not yield significant improvements. This result can be attributed to KD's impact on model confidence; models trained with KD tend to produce more confident predictions than models trained without KD, see Figure~\ref{fig:density}, which is key to degenerate performance on OOD~\citep{utama-etal-2020-towards,sanh2020learning}. Such overconfidence could be a critical factor in degraded performance on OOD tasks.
Moreover, such minimal contribution of KD remains unchanged even when stronger external knowledge (a larger teacher) or a more capable learner (a larger student) is used, see Figure~\ref{fig:agreement}. 

\begin{figure}[t]
\centering
\includegraphics[width=0.497\textwidth]{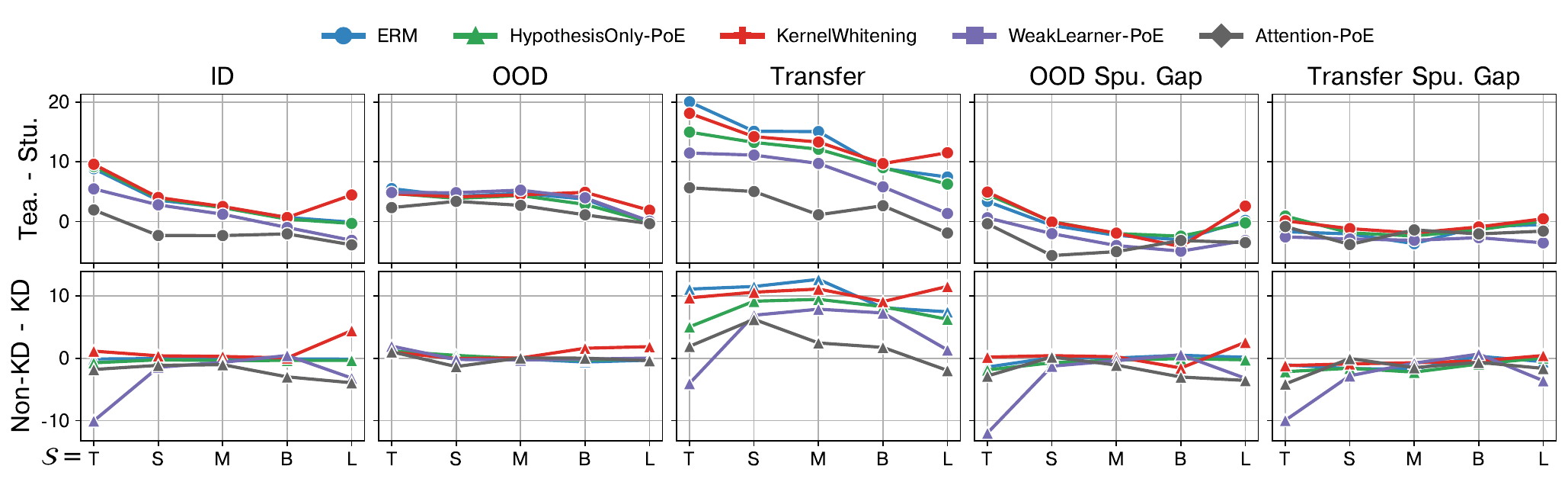}
\caption{\textbf{Comparison Between Debiasing Methods}: performance gap between Teacher and Student (Above), Non-KD and KD (Lower) on text datasets. Detailed results are shown in Appendix.}
\label{fig:lineplot_per_method}
\end{figure}

\subsection{Distillability Across Methods and Backbones} \label{sec:c3}

\paragraph{Different transfer patterns across methods}
Our results show that the transfer patterns are heavily influenced by the formulation of debiasing method. For example, logit-based PoE methods, such as \ETE and \WL, show  similar trends in performance changes and spurious gaps, in contrast to the representation disentanglement method (\KW), see Figure~\ref{fig:lineplot_per_method}.

\paragraph{Sensitivity to backbones}
The distillability of KD appears to varies with the architecture of the backbones and randomness in the training. \KW and \WL are two methods particularly sensitive to the scale of backbone and random seeds, which controls factors such as data sampling and ordering.

\paragraph{Robustness to different biases transfer differently}
We observe that OOD and Transfer show different transfer patterns, where performance gap on Transfer exhibit much larger variations the student scales up, see Figure~\ref{fig:lineplot_per_method}. This suggests that smaller students may outperform larger ones on OOD, indicating that during KD, larger students may become more prone to certain biases (OOD) but more resilient to others.


\paragraph{Universal transfer patterns in debiasing methods}
A number of debiasing methods show consistent changes in robustness after KD, which suggest the potential for an empirical universal transfer pattern. Specifically on text datasets, the performance gap between teacher and student models on OOD and Transfer Spurious Gap fall in the range of [0, 5] and [-5, 0] respectively, see Figure~\ref{fig:lineplot_per_method}. Such change in performance is consistent across different scales of \teascale and \stuscale, which allows for predictable performance after KD. Similarly, the performance gap between models trained using Non-KD and KD remains stable on OOD, falling in range [-1, 1] across different scales.

\begin{figure}[t]
    \centering
    \includegraphics[width=0.49\textwidth]{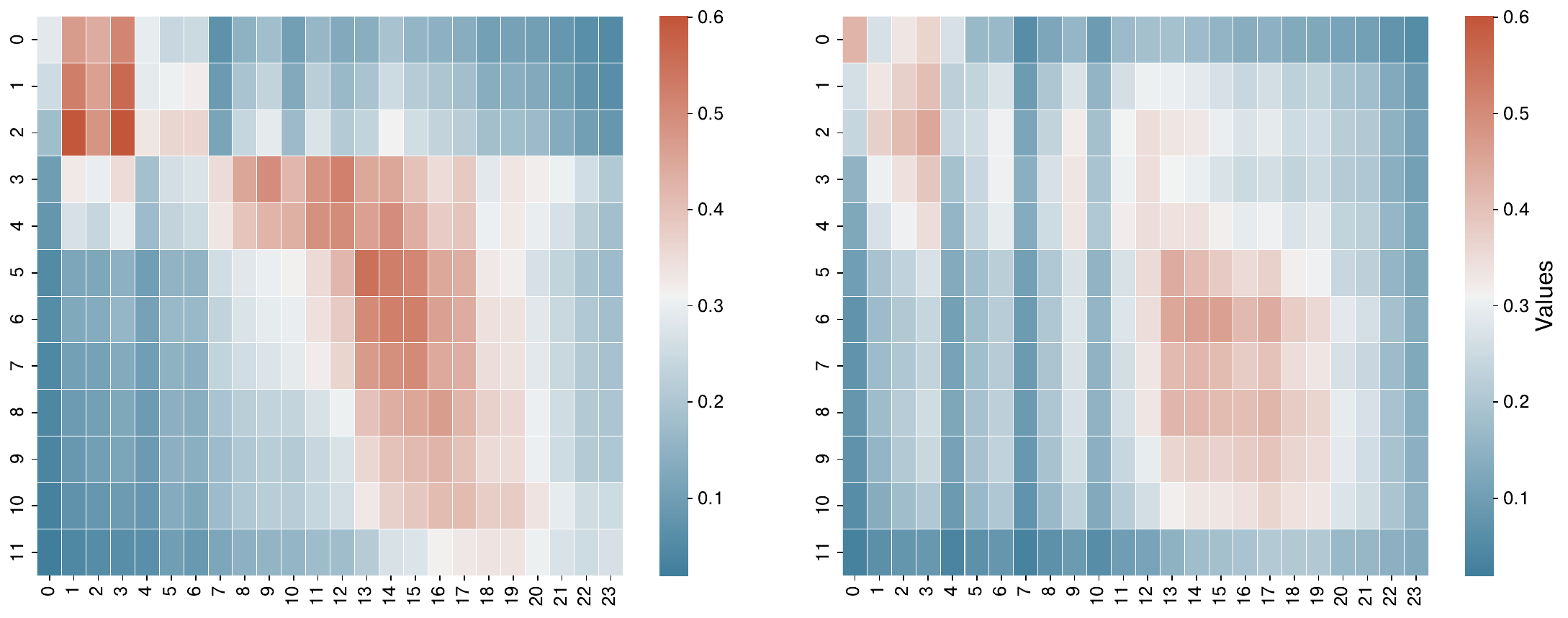}
    \caption{\textbf{C1: Teacher vs. Student}: Centered Kernel Alignment on ID (left) and OOD (right). Highers values indicate higher similarity. X-axis and Y-axis refer to the layers of teacher (\teascale) and student (\stuscale) respectively. See Appendix~\ref{sec:additional_results} for detailed results.}
    \label{fig:cka_tvs}
\end{figure}

\section{Internal Mechanism of Lack of Distillability}

\subsection{Attention}
Attention plays a critical role in making predictions, and biased models may learn spurious attention patterns~\citep{wang-etal-2023-robust}. We hypothesize that the divergence of teacher and student on OOD samples is due to the difference in their internal attention patterns.

Figure~\ref{fig:cka_tvs} shows the difference between internal representations between teacher and student when making predictions on ID and OOD data. After distillation, students try to mimic teachers on ID (left). The earlier layers of students follow earlier layers of teachers, and similarly mid and later layers. This indicates that KD can transfer knowledge of ID data from the larger teachers into smaller students. On OOD (right), however, we observe similar pattern but it is not fully preserved. In particular, it is challenging for the mid and later layers of the students to follow closely to those of the teachers, which explains the performance degradation on OOD after KD, see Figure~\ref{fig:cka_tvs}.

\begin{figure*}[t]
    \centering
    \begin{subfigure}[t]{0.49\textwidth}
    \centering
    \includegraphics[width=\textwidth]{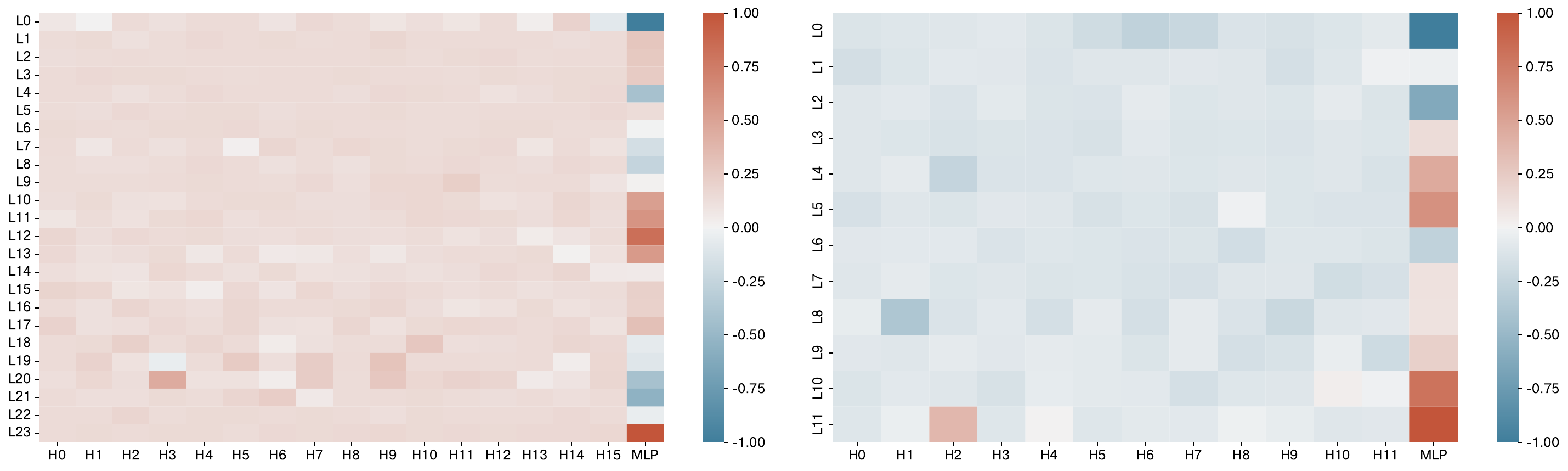}
    \caption{C1: Teacher (Left) vs. Student (Right)}
    \label{fig:circuit_tvs}
    \end{subfigure}
    \hfill
    \begin{subfigure}[t]{0.49\textwidth}
    \centering
        \includegraphics[width=\textwidth]{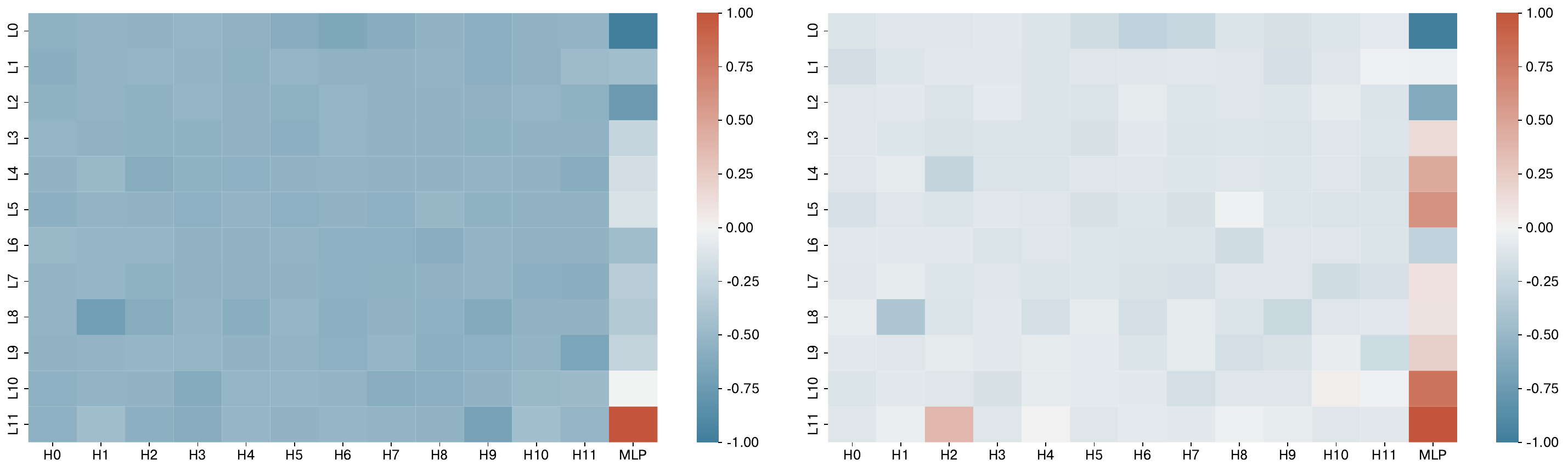}
    \caption{C2: Non-KD (Left) vs. KD (Right)}
    \label{fig:circuit_kvn}
    \end{subfigure}
    \caption{\textbf{Comparison of Discovered Circuit}. L, H, MLP denotes the layer, attention head and MLP components respectively. Each cell shows the causal effect on logit.}
    \label{fig:circuit}
\end{figure*}

\subsection{Circuit}

\paragraph{Teacher vs. Student}
For teacher models, we observe a consistently moderate positive effect across all attention heads. While the effect of MLP layers can be negative. However, knowledge distillation have reverted the effect of attention heads into negative. There is not a unified pattern on the MLP layers.

\paragraph{Non-KD vs. KD}
On Non-KD trained models, attention heads across all layers have negative effect on the final logit, except for the final MLP layer which has a strong positive effect. On the contrary, the KD trained models tend to emphasize MLP layers and suppress the contribution on attention heads. This is likely due to learning process of KD is centered on matching logits, which may miss some contributions of the earlier layers.

\section{Potential Solutions}

Based on the above analyses, we summarize the key findings on distilling debiasing models as follows:
\begin{itemize}
\itemsep0pt
    \item Training distribution significantly affect successful distillation of debiasing capabilities.
    \item Student models with similar scale to their teachers can better obtain debiasing knowledge from their teachers.
    \item The objectives of KD may introduce additional optimization challenges, especially with the presence of debiasing objectives.
\end{itemize}

To further improve the distillability of debiasing methods, we propose three solutions:

\paragraph{Data augmentation (DA)}
There is broad evidence that models becomes biased by relying on spurious features in the training set~\citep{wu-etal-2022-generating,ahn2023mitigating}, which is amplified by misrepresentation of specific classes or labeling errors. Prior studies have highlighted the important role of data in knowledge distillation~\citep{stanton2021does}. Based on these prior studies and our findings, we hypothesize that providing high quality data and augmenting data size can improve the process of distilling the debiasing capability from teacher to student. For text datasets, we employ the data generated by Seq-Z filtering~\citep{wu-etal-2022-generating} as training set for both teacher and student models. For image datasets, we employ training and validation sets where the sub-groups are equally represented~\citep{kirichenko2023last}.\looseness-1

\paragraph{Iterative knowledge distillation (IKD)}
Our results indicate that the transfer of debiasing capabilities is more effective between teachers and students of similar scales. Therefore, we propose to leverage Iterative Knowledge Distillation (IKD)~\citep{Liu_Wang_Shang_Wu_2023}: given a teacher of scale $\mathcal{S}_N$, we first distill it to a student of scale $\mathcal{S}_{N-1}$, where $\mathcal{S}_{N-1}$ is the closest neighbor of $\mathcal{S}_N$ in scale. Then the newly-distilled student acts as a teacher and transfer the knowledge to a model of smaller scale $\mathcal{S}_{N-2}$, where $\mathcal{S}_{N-2}$ is the closest neighbor of $\mathcal{S}_{N-1}$ in scale. We repeat this process iteratively by gradually decreasing student scale, such that the knowledge can be transferred smoothly from a large scale model to a small scale model. This step-wise approach enables a smooth knowledge transfer from larger to smaller scale models, and potentially improves debiasing effectiveness at each step.

\paragraph{Initialize student with teacher weights (Init)}
Previous research by~\citet{stanton2021does} has discovered that initializing a student model with the weights of its teacher can increase their centered kernel alignment~\citep{cka} in activation space. This approach can head-start the student model with a stronger debiasing capability from the teacher. It can also help alleviate potential optimization obstacles and stuck in local optima. If the teacher and student models are of the same scale, we initialize the student with the teacher parameters. If the teacher is larger, we initialize the student with the first few layers of the teacher.

\paragraph{Results}
Table~\ref{tab:solution_res} shows that all three solutions result in improved distillability. Specifically, on spurious gap between teacher and student, data augmentation (DA), iterative knowledge distillation (IKD), and initialization with teacher weights (Init) yield performance gains of 4.5, 2.8, 1.2 absolute points compared to vanilla KD across datasets and backbones respectively. On spurious gap between Non-KD and KD, DA, IKD, and Init outperforms vanilla KD by 1.7, 0.6 and 0.2 absolute points respectively.
We find that DA achieves the largest improvement, since the root cause of spurious correlations come from the underlying dataset~\citep{chen2018my}. Debiasing the dataset itself can benefit all training methods including knowledge distillation. 
As noted by previous work~\citep{stanton2021does}, Init may facilitate teacher-student agreement in activation space, but result in non-significant gains, which aligns with our findings as well.

\begin{table}[t]
\caption{Improvement of distillability. Vanilla refers to standard knowledge distillation, DA, IKD, and Init represent data augmentation, iterative knowledge distillation, and initialization of student with teacher weights (Init) as our three solutions to improve the distillability of debiasing methods.}
\label{tab:solution_res}
\tiny
\setlength{\tabcolsep}{3pt}
\centering
    \begin{tabular}{c|ccc}
    \toprule
    Difference in & ID ($\downarrow$) & OOD ($\downarrow$) & Spu. Gap ($\downarrow$) \\
    \midrule
    \multicolumn{4}{l}{Teacher - Student} \\
    \midrule
    Vanilla & 5.1 & 7.3 & 12.7 \\
    + DA    & 2.3 & 5.4 &  8.2 \\
    + IKD   & 3.6 & 5.9 &  9.9 \\
    + Init. & 4.7 & 6.5 & 11.5 \\
    \midrule
    \multicolumn{4}{l}{Non KD - KD} \\
    \midrule
    Vanilla & 1.4 & 0.7 & 2.2 \\
    + DA    & 0.2 & 0.2 & 0.5 \\
    + IKD   & 1.0 & 0.5 & 1.6 \\
    + Init. & 1.3 & 0.7 & 2.0 \\
    \bottomrule
    \end{tabular}


\end{table}

\section{Conclusion}
We present the first study on the distillability of debiasing capabilities between neural models, and the extent of bias transfer through knowledge distillation (KD). We evaluate eight popular debiasing methods and five scales of backbones on four datasets. Extensive experiments show that 
vanilla KD does not consistently preserve debiasing capabilities; in many cases, student models become more reliant on spurious correlations than their teachers; 
the effectiveness of debiasing transfer depends on model scale similarity--distillation works best when teacher and student models are comparable in complexity; and 
larger teachers do not always yield more robust students, which indicates the need for targeted debiasing strategies in KD. 
We propose three solutions--data augmentation, iterative KD, and student initialization--which significantly improve the distillability of debiasing methods and contribution of KD on debiasing. \looseness-1

In future we will investigate self-distilled debiasing, where the student iteratively distills knowledge from itself rather than relying on a fixed teacher. A potential improvement is to explicitly guide the student using counterfactual data augmentation during distillation. 


\section*{Limitations}
Despite delivering significant amount of discoveries, our work has certain limitations. 
Firstly, our experiments are mainly conducted on logit-based knowledge distillation. The effect of other knowledge distillation methods has not been explored. 
Secondly, the work does not explore the scenario where multiple teachers participate in the distillation process.

\section*{Ethical Considerations}
Our research focuses on mitigating dataset biases in text and vision datasets, and understanding why debiasing methods may fail under knowledge distillation.
The broader impacts of our work are in advancing dataset fairness and potentially enhancing decision-making based on data. Our work contributes to improving the accuracy and reliability of NLP and vision models, as well as their trust and adoption. 



\bibliography{anthology,custom}
\nocite{cheng2024mu}
\nocite{chen2025frog}

\newpage
\appendix
\section{Related Work}\label{sec:related}

\paragraph{Bias mitigation in NLU}
Debiasing approaches usually employ a biased model to inform the training of a robust model~\citep{clark-etal-2019-dont,karimi-mahabadi-etal-2020-end,sanh2020learning,utama-etal-2020-towards,cheng24c_interspeech}. Other methods aim at learning debiased or robust representations~\citep{gao-etal-2022-kernel,lyu2023feature,wang-etal-2023-robust,jeon-etal-2023-improving,reif-schwartz-2023-fighting}, or removing bias-encoding parameters~\citep{meissner-etal-2022-debiasing,yu-etal-2023-unlearning,cheng-amiri-2025-equalizeir}. Other works include measurement of bias of specific words with statistical test~\citep{gardner-etal-2021-competency}, generating non-biased samples~\citep{wu-etal-2022-generating}, identification of bias-encoding parameters~\citep{yu-etal-2023-unlearning}, when bias mitigation works~\citep{ravichander-etal-2023-bias}, bias transfer from other models~\citep{jin-etal-2021-transferability}, and bias removal with model unlearning~\citep{cheng2024mu,chen2025frog}.

\paragraph{Bias mitigation in vision}
In vision, worst-group performance is measured as a sign of model robustness. Several works investigates how to learn debiased models from failure cases~\citep{lff}, 
biased representations~\citep{pmlr-v119-bahng20a}, 
multiple biased models~\citep{lwbc}, and 
by simply re-training the last layer of a neural model (i.e. the classification layer) with additional equally represented data~\citep{kirichenko2023last,last_layer_fewer}. \citet{Li_2023_CVPR} showed that multiple spurious features can occur in a dataset, while suppressing one may inevitably boost another one. Other perspectives for debiasing include causal attention~\citep{Wang_2021_ICCV}, building uniform margin classifiers~\citep{NEURIPS2023_e3546030}, using representations from earlier layers~\citep{Tiwari_2024_WACV}, and neural collapse~\citep{Wang_2024_CVPR}, where feature space collapses into a stable geometric structure that results in robustness and generalizability.

\paragraph{Knowledge distillation}
Knowledge Distillation (KD) is initially proposed to transfer knowledge from a larger model (teacher) to a smaller model (student), by encouraging the student to follow the teacher on prediction logits~\citep{hinton2015distilling}, learned features~\citep{romero2014fitnets,10.1007/978-3-030-58586-0_20}, attention map~\citep{zagoruyko2017paying,Chen_Mei_Zhang_Wang_Wang_Feng_Chen_2021}, activation patterns~\citep{huang2017like,Heo_Lee_Yun_Choi_2019}. 
Later works discovered that KD can be viewed as a special form of regularization similar to label smoothing~\citep{Szegedy_2016_CVPR}, providing no task-specific knowledge. However, on text classification tasks, whether KD can regularize the student depends on the choice of teacher model~\citep{sultan-2023-knowledge}, which may result in opposite model confidence between teacher and student compared to label smoothing.
\citet{stanton2021does} discovers that optimization and dataset details are crucial to matching students to teachers, and such matching does not guarantee better generalization ability of students.
\citet{xue2023the} investigates cross-modal KD, where the teacher functions on a different modality or extra modalities than student. The authors proposes modality fusing hypothesis, which claims that modality decisive features are critical for the effectiveness of cross-modal KD.
However, despite briefly discussed~\citep{Cho_2019_ICCV,Tiwari_2024_WACV}, the potential of knowledge distillation to transfer debiasing capabilities across different modalities and backbone models remains 
underexplored and poorly understood in existing work.\looseness-1

\section{Details of Dataset}\label{sec:dataset}

We describe the details of each dataset below:

\begin{itemize}
    \item {\bf CelebA~\citep{celeba}} consists of 16k images of celebrity faces, where the objective is to predict ``Blond\_Hair'' given ``Male'' as a spurious attribution.
    
    \item {\bf Waterbird~\citep{SagawaDistributionally}} consists of synthetic images of birds from CUB dataset~\citep{WahCUB_200_2011} and backgrounds (land \& water) from Places~\citep{7968387} dataset. The objective is to correctly infer ``land bird'' or ``water bird,'' given the background as misleading information. 
    
    \item {\bf MNLI~\citep{williams-etal-2018-broad}} consists of 39k natural language inference (NLI) samples from various domains, where the objective is to classify relationship between a premise and a hypothesis as ``Entailment'', ``Contradiction'', or ``Neutral''. Previous studies discover that models are prone to negation words, lexical overlap, and sub-sequence biases in NLI task~\citep{naik-etal-2018-stress,mendelson-belinkov-2021-debiasing}. We use HANS~\citep{mccoy-etal-2019-right} as the out-of-distribution test set (OOD) and SNLI~\citep{bowman-etal-2015-large} as the transfer test set (Transfer), detailed below.
    
    \item {\bf QQP~\citep{sharma2019natural}} is a paraphrase identification (PI) dataset with 43k samples, where the objective is to predict if two questions are paraphrases of each other. Similar to MNLI, models are likely to be mislead by lexical overlap between two questions. We exploit PAWS~\citep{zhang-etal-2019-paws} as the out-of-distribution test set (OOD) and MRPC~\citep{dolan-brockett-2005-automatically} as the transfer test set (Transfer), detailed below.

\end{itemize}

\section{Details on Debiasing Methods}\label{sec:methods}
Experiments are conducted on a comprehensive list of commonly used debiasing methods, each of which is designed with special formulation and assumptions. 

\begin{itemize}
\item {\bf Empirical Risk Minimization (ERM)} is the standard training method that minimizes the empirical risk on a dataset. This is akin to fine-tuning a pre-trained model on a dataset using cross-entropy loss with no debiasing strategy, which works for both image and text datasets.

\item {\bf \ETE}~\citep{karimi-mahabadi-etal-2020-end} assumes the hypothesis part of NLI datasets contains biases. It trains a hypothesis-only (biased) model to measure the bias of each sample, and uses Product-of-Experts (PoE)~\citep{10.1162/089976602760128018} to adjust the confidence of the debiased model according to the confidence of the biased model. This approach is evaluated on text datasets.

\item {\bf \WL}~\citep{sanh2020learning} leverages weak learners to capture and model bias, including bias of unknown type. It trains a 2-layer BERT as a biased model and exploits PoE to train the debiased model. This approach is evaluated on text datasets. 

\item {\bf \KW}~\citep{gao-etal-2022-kernel} aims at learning isotropic sentence embeddings with disentangled robust and spurious representations, with Nystr\"{o}m kernel~\citep{Xu_Jin_Shen_Zhu_2015}. This approach is evaluated on text datasets.

\item {\bf \READ}~\citep{wang-etal-2023-robust} assumes that the attention to \texttt{[CLS]} token in text classification is biased and introduces PoE on attention weights to learn robust attention patterns for bias mitigation. This approach is evaluated on text datasets.

\item {\bf \DAMP}~\citep{NEURIPS2023_e3546030} assuming the standard cross-entropy loss encourages models to prioritize shortcuts over robust features, this model proposes to scale the loss by a temperature. This approach is evaluated on image datasets.

\item {\bf \DFR}~\citep{kirichenko2023last} discovers that simply retraining the last layer of a neural model--the classification layer in supervised tasks--on top of the existing biased feature extractor is good strategy for bias mitigation. This approach is evaluated on image datasets.

\item {\bf \PGD}~\citep{ahn2023mitigating} trains a debiased model with non-uniform sampling probability, obtained from per-sample gradient norm of a biased model. This approach is evaluated on image datasets. 

\end{itemize}

\section{Implementation details}\label{sec:implementation}
We follow previous debiasing works for implementation details.  For text datasets, we train each debiasing method with Adam optimizer, learning rate $5e-5$, 5 epochs, both KD and Non-KD. For image datasets, we train each debiasing method with Adam optimizer, learning rate $4e-5$, 100 epochs, both KD and Non-KD. For all other hyperparameters, we follow each debiasing method's best-performing setting.

We show the details of backbone models in Table~\ref{tab:backbone}.

\begin{table}[b]
\setlength{\tabcolsep}{3.6pt}
\caption{Different scales of backbones in our experiments. $h$ and $d$ denote number of hidden layers and size of hidden dimension respectively. T, S, M, B, L refer to Tiny, Small, Medium, Base and Large version of the backbone. See Appendix for more details.}
\label{tab:backbone}
\centering
\small
\begin{tabular}{l|cc|cc||cc|cc}
\toprule
    \multirow{2}{*}{Scale} & \multicolumn{2}{c}{BERT} & \multicolumn{2}{c||}{T5} & \multicolumn{2}{c}{ResNet} & \multicolumn{2}{c}{ViT}\\
    \cmidrule(lr){2-3} \cmidrule(lr){4-5} \cmidrule(lr){6-7} \cmidrule(lr){8-9}
    & $h$ & $d$ & $h$ & $d$ & $h$ & $d$ & $h$ & $d$ \\
    \midrule
    T &  2 &  128 &  4 &  256 &  18 &  512 & 12 &  192 \\
    S &  4 &  256 &  8 &  384 &  34 &  512 & 12 &  384 \\
    M &  8 &  512 & 16 &  512 &  50 & 2048 & 12 &  768 \\
    B & 12 &  768 & 24 &  768 & 101 & 2048 & 24 & 1024 \\
    L & 24 & 1024 & 48 & 1024 & 152 & 2048 & 32 & 1280 \\
    \bottomrule
    \end{tabular}
\end{table}

\section{Results on Image Datasets}
On image datasets, we observe similar results on text datasets. Specifically, we see that KD fall short on distilling the debiasing capabilities. Such ability is transferred more smoothly as teacher and student get similar in scale.

\begin{figure*}[t]
    \vskip 0.2in\centering
    \includegraphics[width=0.95\textwidth]{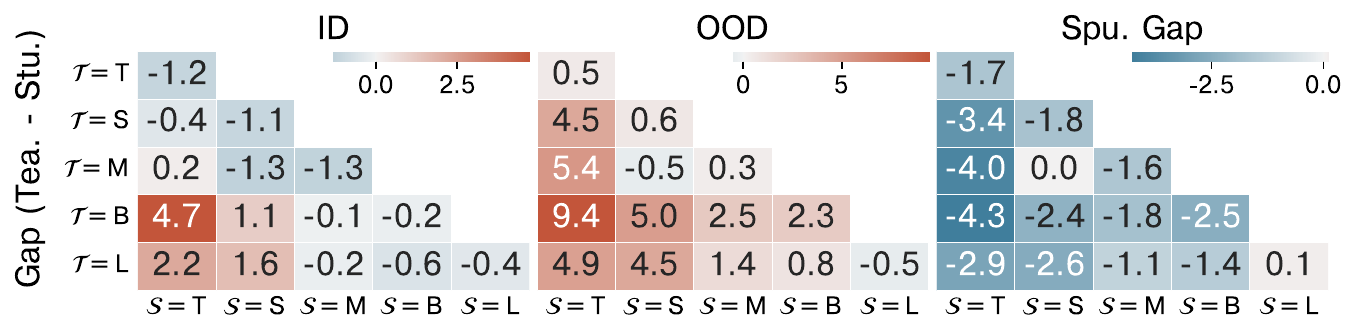}
    \caption{\textbf{C1: Teacher vs. Student}: average performance gaps between teacher and student models on ID, OOD, and Spurious Gap across image datasets. X-axis and Y-axis show the scale of student (\stuscale) and teacher (\teascale) respectively. Each cell shows the performance gap between corresponding scales of a teacher and a student.}
    \label{fig:heatmap_tvs_image}
\end{figure*}

\begin{figure*}[t]
    \vskip 0.2in\centering
    \includegraphics[width=0.95\textwidth]{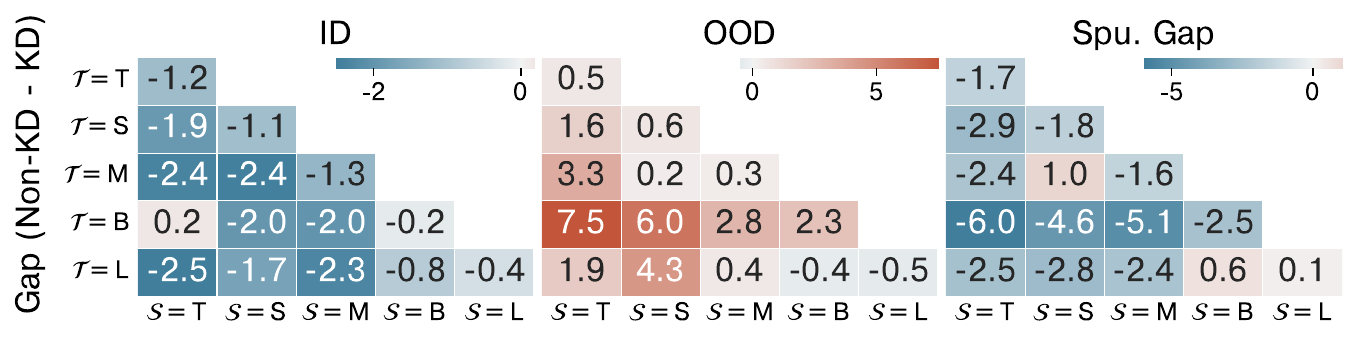}
    \caption{\textbf{C2: Non-KD vs. KD}: average performance gaps between Non-KD and KD models on ID, OOD, and Spurious Gap across image datasets. X-axis and Y-axis show the scale of student (\stuscale) and teacher (\teascale) respectively. Each cell shows the performance gap between corresponding scales of a teacher and a student.}
    \label{fig:heatmap_kvn_image}
\end{figure*}

\section{Detailed Results on Debiasing Methods and Backbones}\label{sec:additional_results}
We present the detailed results of individual debiasing method and backbone below.

\begin{figure}[t]
    \vskip 0.2in
    \centering
    \includegraphics[width=0.49\textwidth]{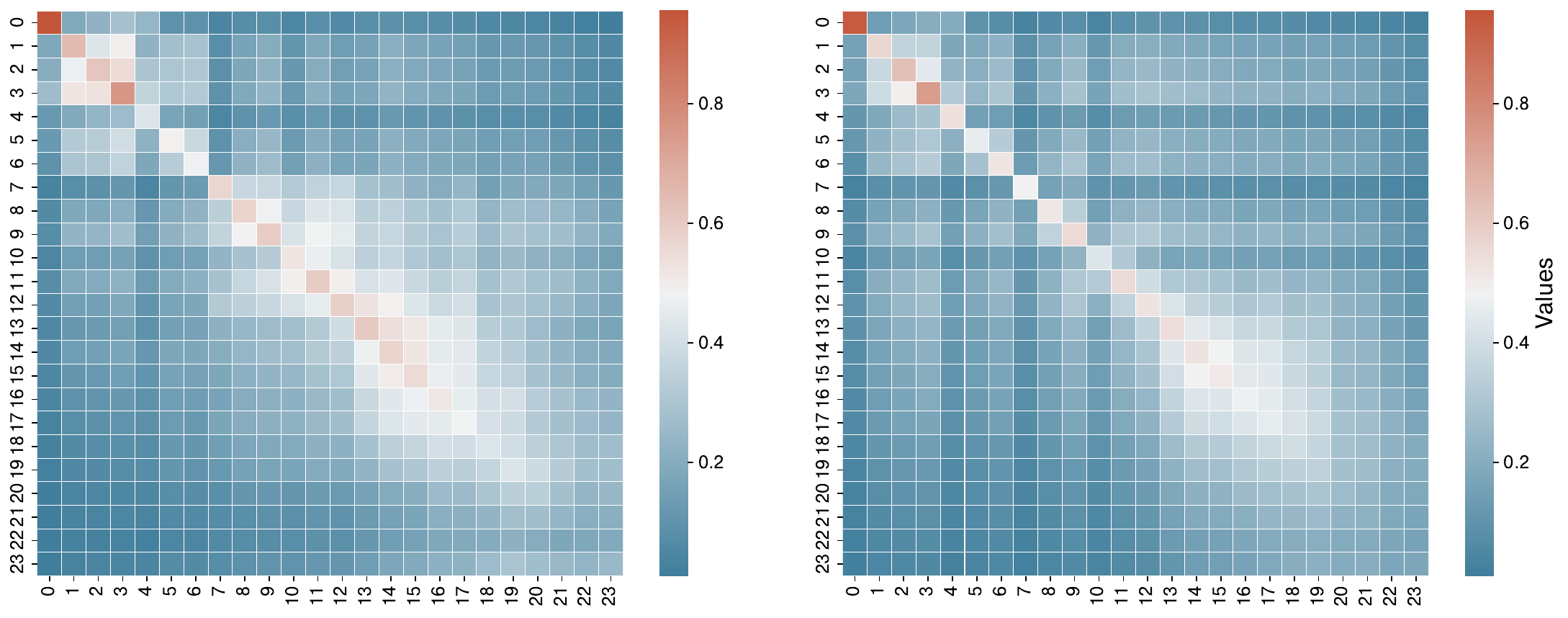}
    \caption{\textbf{C2: KD vs. Non-KD}: Centralized Kernel Alignment. Highers values indicate higher similarity. X-axis and Y-axis refer to the layers of KD (\stuscale) and Non-KD (\studentnonkd) respectively.}
    \label{fig:cka_kvn}
\vskip -0.2in
\end{figure}

\begin{figure*}[t]
    \vskip 0.2in\centering
    \includegraphics[width=0.95\textwidth]{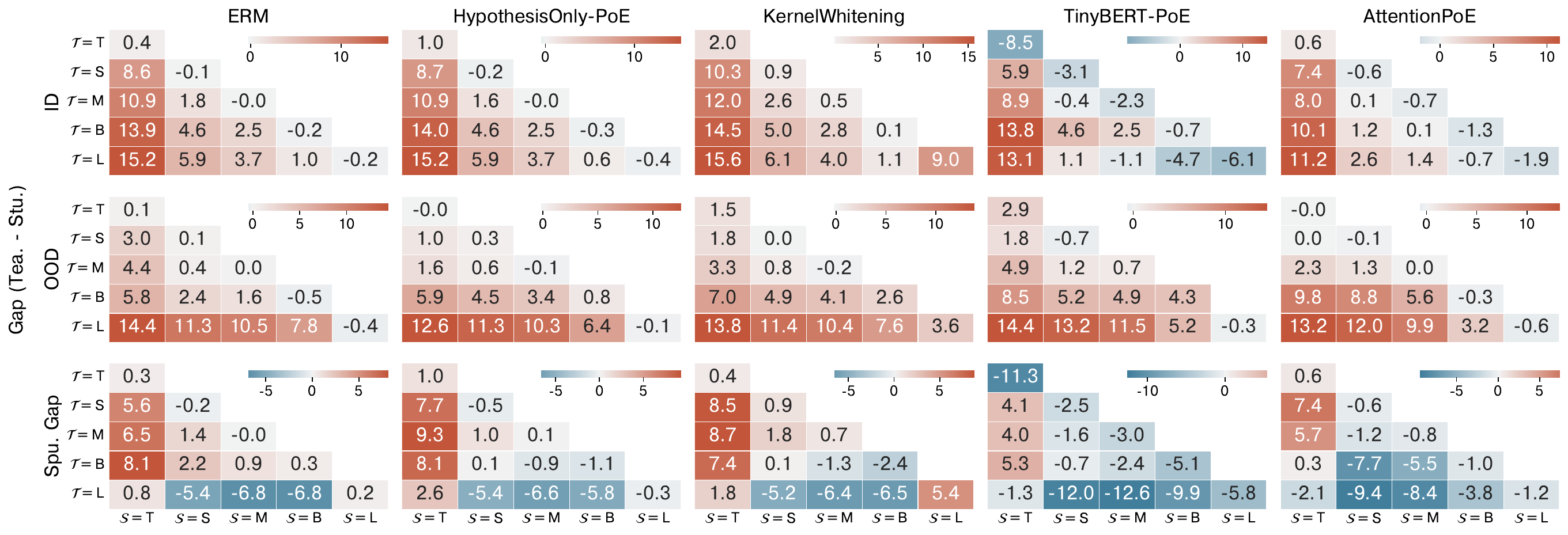}
    \caption{\textbf{C1: Teacher vs. Student}: average performance gaps between teacher and student models on ID, OOD, and Spurious Gap on BERT.}
    \label{fig:heatmap_per_method_tvs_bert}
\end{figure*}

\begin{figure*}[t]
    \vskip 0.2in\centering
    \includegraphics[width=0.95\textwidth]{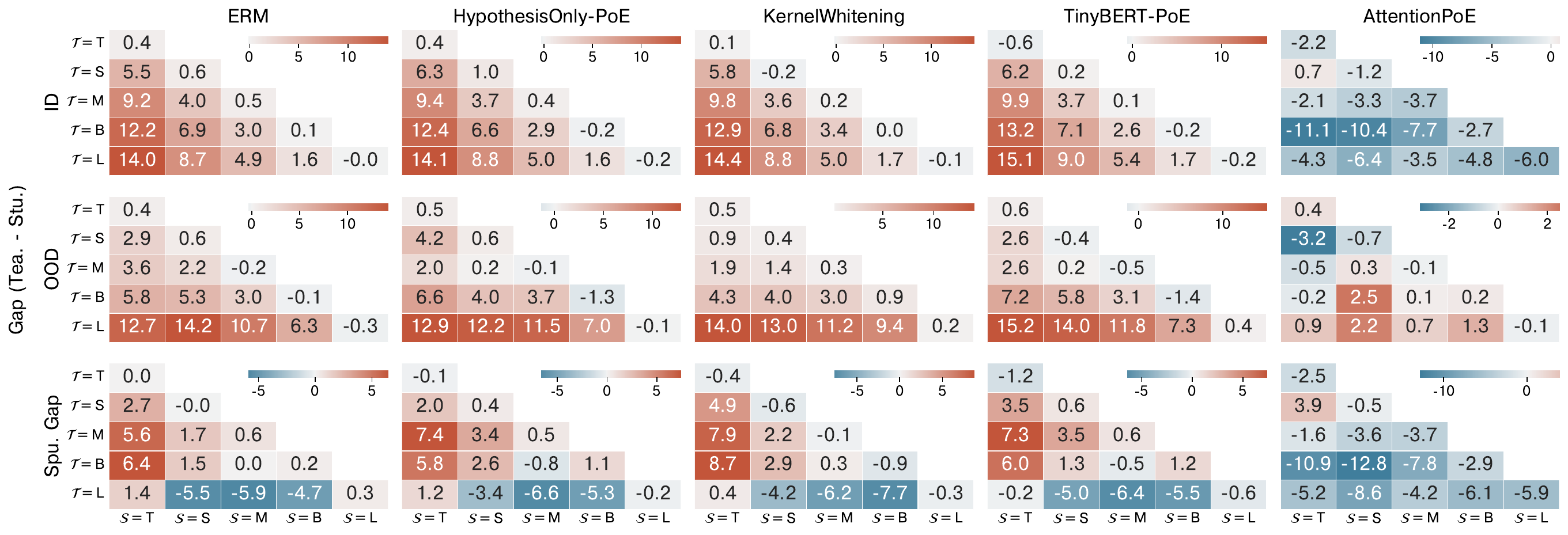}
    \caption{\textbf{C1: Teacher vs. Student}: average performance gaps between teacher and student models on ID, OOD, and Spurious Gap on T5.}
    \label{fig:heatmap_per_method_tvs_t5}
\end{figure*}

\begin{figure*}[t]
    \vskip 0.2in\centering
    \includegraphics[width=0.95\textwidth]{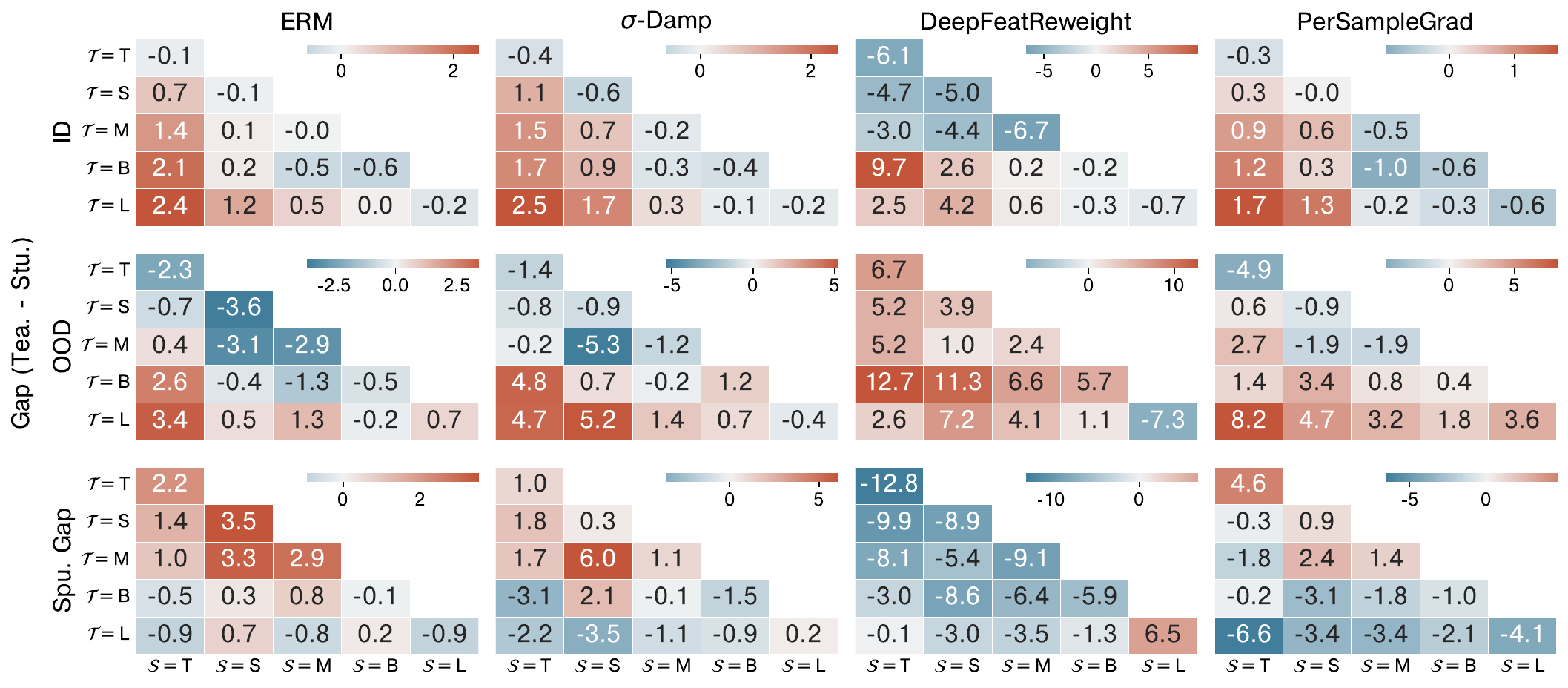}
    \caption{\textbf{C1: Teacher vs. Student}: average performance gaps between teacher and student models on ID, OOD, and Spurious Gap on ResNet.}
    \label{fig:heatmap_per_method_tvs_resnet}
\end{figure*}

\begin{figure*}[t]
    \vskip 0.2in
    \centering
    \includegraphics[width=0.95\textwidth]{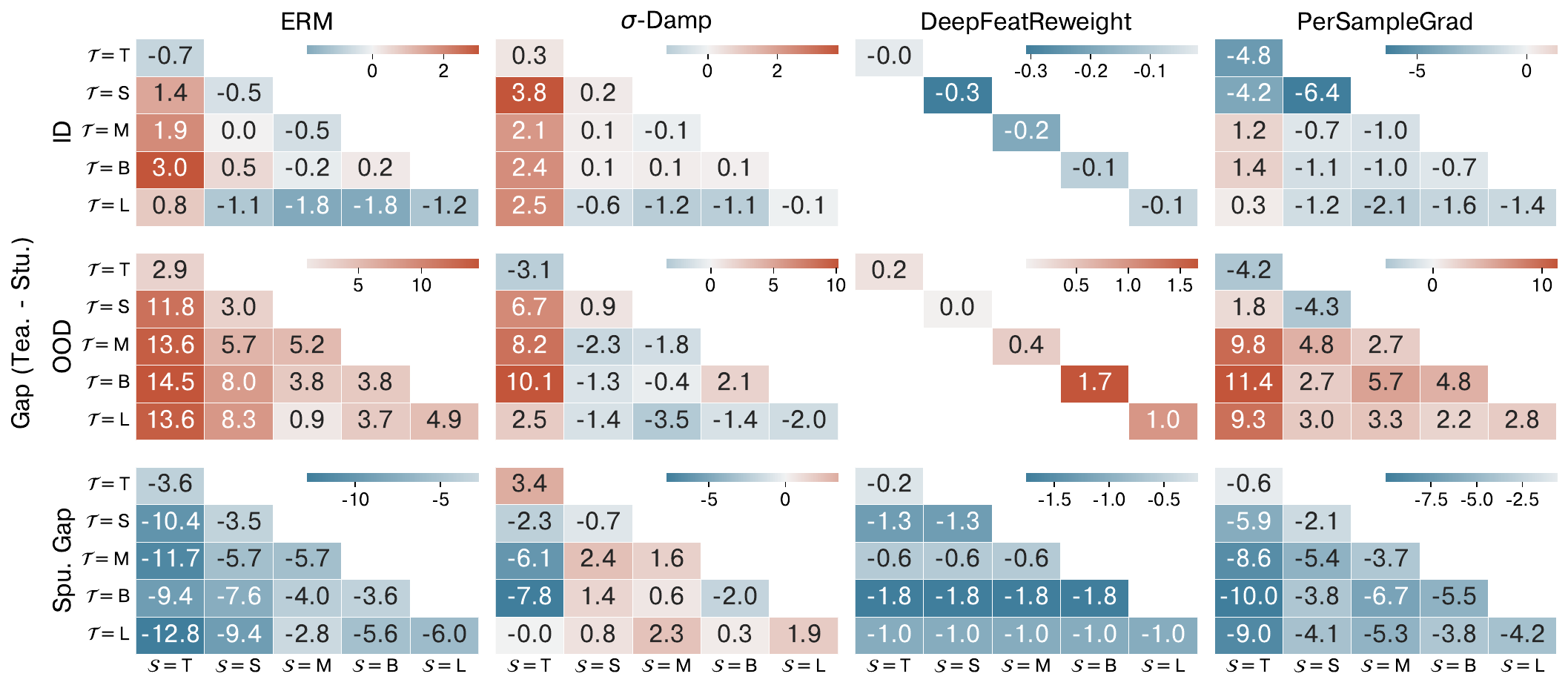}
    \caption{\textbf{C1: Teacher vs. Student}: average performance gaps between teacher and student models on ID, OOD, and Spurious Gap on ViT.}
    \label{fig:heatmap_per_method_tvs_vit}
\end{figure*}

\begin{figure*}[t]
    \vskip 0.2in\centering
    \includegraphics[width=0.95\textwidth]{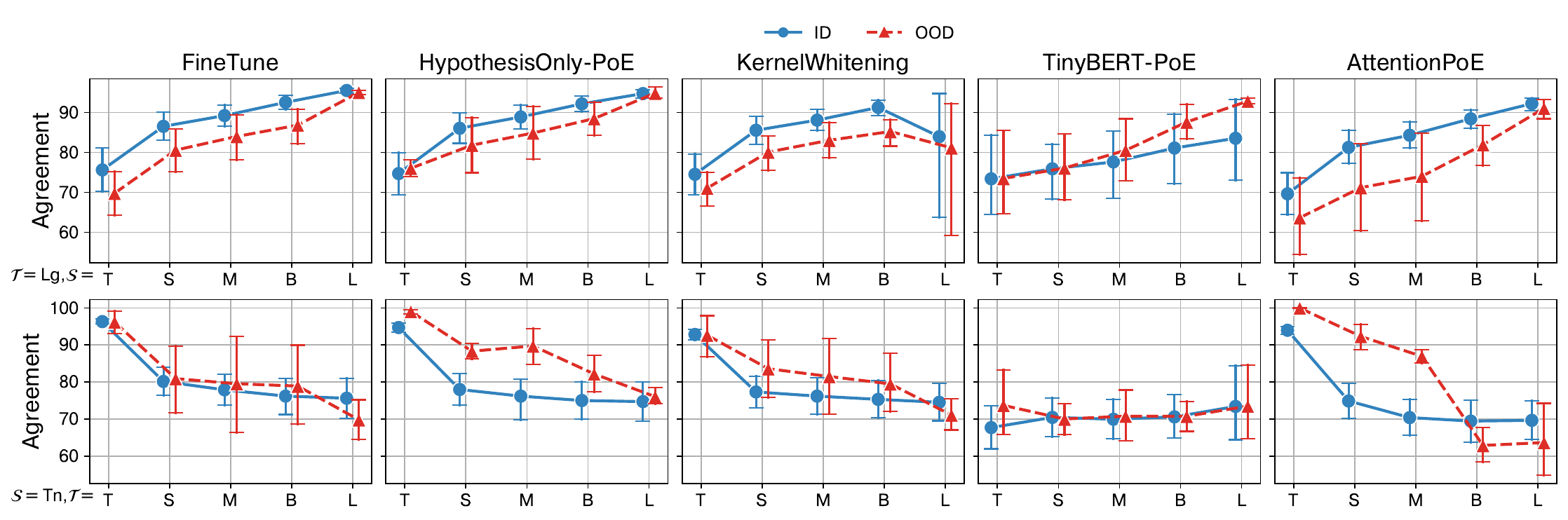}
    \caption{\textbf{C1: Teacher vs. Student}: prediction agreement on BERT. Left: varying \stuscale (X-axis) given a fixed teacher with \teascale = L. Right: varying \teascale (X-axis) given a fixed student with \teascale = T. Agreement increases as the scale of teacher and student get closer.}
    \label{fig:agreement_per_method_tvs_bert}
\end{figure*}


\end{document}